\documentclass[runningheads]{llncs}

\usepackage[T1]{fontenc}
\usepackage{graphicx}
\usepackage{graphicx}
\usepackage{booktabs}
\usepackage{amsmath}
\usepackage{amssymb}

\usepackage{cleveref}
\crefname{figure}{Fig.}{Figs.}
\Crefname{figure}{Fig.}{Figs.}
\crefname{table}{Table}{Tables}
\Crefname{table}{Table}{Tables}

\usepackage{cite}

\usepackage{subcaption} 
\usepackage{float}
\usepackage{tabularx}
\usepackage{multirow}
\usepackage{soul}
\usepackage{array}
\usepackage{adjustbox}
\usepackage{caption}
\usepackage{color,xcolor}
\usepackage{colortbl}
\definecolor{mygray}{gray}{.9}
\newcommand{\eg}{e.g., }

\begin{document}

\title{AllWeather-Net: Unified Image Enhancement for Autonomous Driving Under Adverse Weather\\ and Low-Light Conditions}

\author{Chenghao Qian\inst{1}\orcidID{0009-0008-7054-9556} \and
Mahdi Rezaei \inst{1}\orcidID{0000-0003-3892-421X}      \and
Saeed Anwar \inst{2}\orcidID{0000-0002-0692-8411}      \and
Wenjing Li\inst{1, *} \orcidID{0000-0003-3201-6675}      \and
Tanveer Hussain \inst{3}\orcidID{0000-0003-4861-8347}      \and
Mohsen Azarmi \inst{1}\orcidID{0000-0003-0737-9204}      \and
Wei Wang \inst{4}\orcidID{0000-0001-8676-1190}}
\authorrunning{C. Qian et al.}
\titlerunning{AllWeather-Net: Unified Image Enhancement for Autonomous Driving}
\institute{University of Leeds, Leeds, UK \and
Australian National University, Canberra, Australia \and
Edge Hill University, Ormskirk, UK\and
Shenzhen Campus of Sun Yat-sen University, ShenZhen, China
\\
*\email{wjli007@mail.ustc.edu.cn}}

\maketitle              
\vspace{-6mm}
\begin{figure}[H]
  \centering
  \begin{minipage}{10cm}  
    \begin{subfigure}{\linewidth}
      \centering
      \includegraphics[width=10cm]{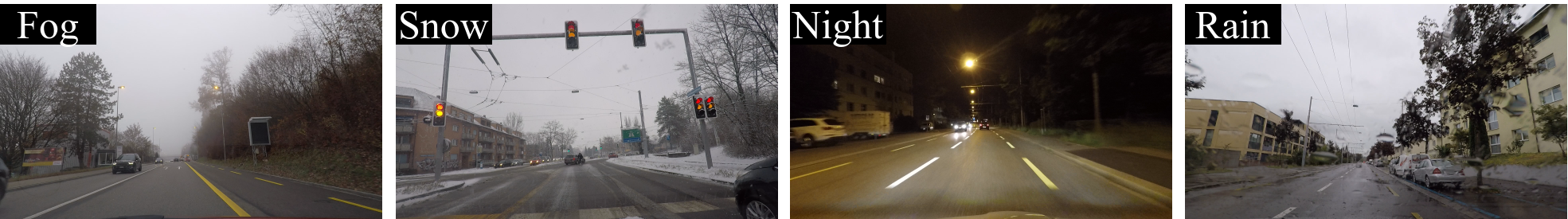}
      \subcaption{Original Input Images}
      \label{fig:input}
    \end{subfigure}

    \begin{subfigure}{\linewidth}
      \centering
      \includegraphics[width=10cm]{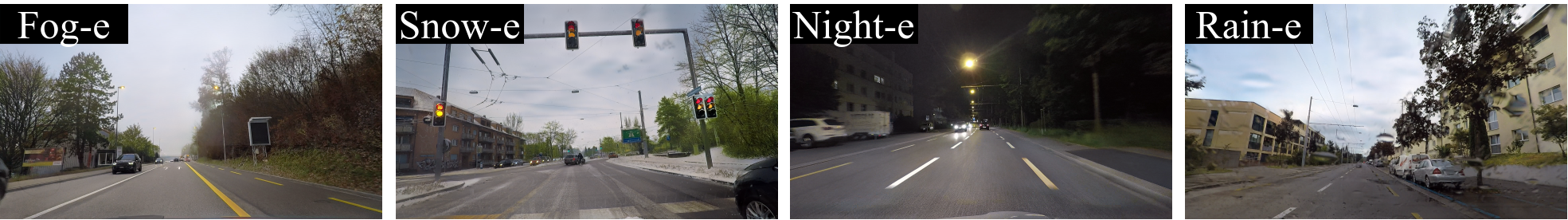}
      \subcaption{Enhanced Output Images}
      \label{fig:output}
    \end{subfigure}
    \vspace{-6mm}
    \caption{\small Given images captured under adverse conditions in (a), we propose a method that can effectively adjust color and texture, modify lighting and shadows, and remove weather effects within a unified model. This results in a visually appealing appearance that resembles normal, day-like weather conditions (b), thereby enhancing the robust performance of autonomous driving perception systems.}
    \label{fig:img}
  \end{minipage}
\end{figure}
\vspace{-10mm}
\begin{abstract}
Adverse conditions like snow, rain, nighttime, and fog, pose challenges for autonomous driving perception systems. 
Existing methods have limited effectiveness in improving essential computer vision tasks, such as semantic segmentation, and often focus on only one specific condition, such as removing rain or translating nighttime images into daytime ones.
To address these limitations, we propose a method to improve the visual quality and clarity degraded by such adverse conditions. 
Our method, AllWeather-Net, utilizes a novel hierarchical architecture to enhance images across all adverse conditions. This architecture incorporates information at three semantic levels: scene, object, and texture, by discriminating patches at each level. 
Furthermore, we introduce a Scaled Illumination-aware Attention Mechanism (SIAM) that guides the learning towards road elements critical for autonomous driving perception. SIAM exhibits robustness, remaining unaffected by changes in weather conditions or environmental scenes. 
AllWeather-Net effectively transforms images into normal weather and daytime scenes, demonstrating superior image enhancement results and subsequently enhancing the performance of semantic segmentation, with up to a 5.3\% improvement in mIoU in the trained domain.  We also show our model's generalization ability by applying it to unseen domains without re-training, achieving up to 3.9 \% mIoU improvement. Code can be accessed at: \textit{https://github.com/Jumponthemoon/AllWeatherNet.}
\end{abstract}
\section{Introduction}
\label{sec:intro}

Autonomous driving systems heavily rely on clear and optimal environmental images; however, these are not guaranteed in real life due to natural conditions, like snow, rain, fog, low light at night, etc. This can significantly reduce visibility and distort the captured information within an image, which impacts the performance of autonomous driving perception systems, including but not limited to object detection and semantic segmentation. 

To counter the mentioned problem, some methods remove weather artifacts via deraining~\cite{yang2019joint,wang2022uformer}, dehazing~\cite{berman2016non,zhang2019joint}, and desnowing~\cite{liu2018desnownet,zhang2021deep,wang2023smartassign}. Moreover, some unified frameworks~\cite{jose_valanarasu_transweather_2022,li_all_nodate-1,Chen2022MultiWeatherRemoval} handle three types of weather while mainly focusing on removing hydrometer particles, neglecting alterations in color and texture details; hence, restricting their effectiveness under adverse weather conditions for autonomous driving computer vision systems.

\begin{figure}[htbp]
        \centering
        \includegraphics[width=\linewidth]{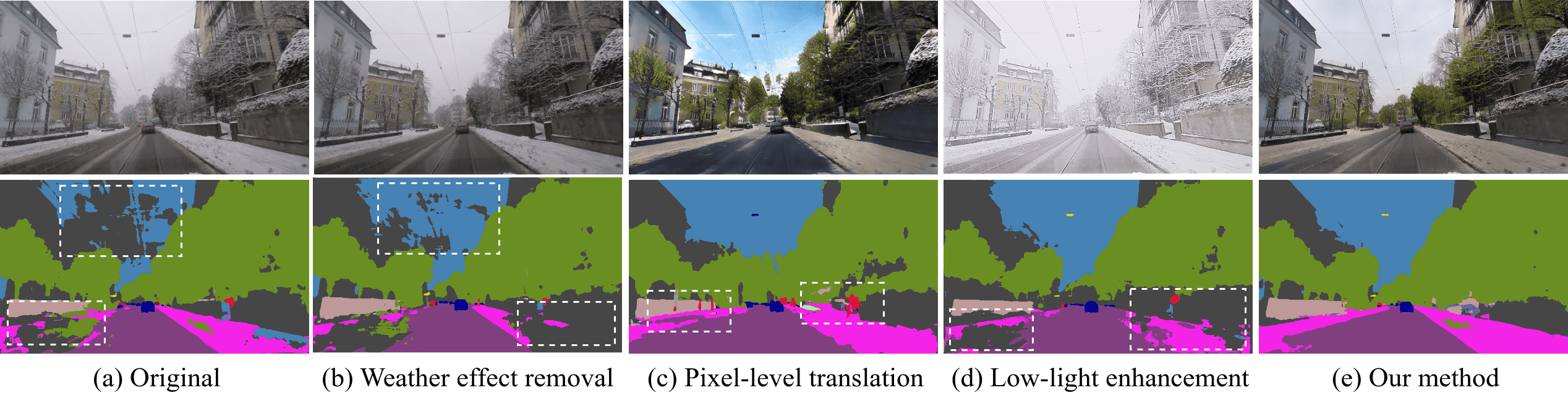}
        \caption{(a) Original Image. The evaluation of image processing techniques for semantic segmentation under adverse conditions reveals the deficiencies of (b) weather effect removal~\cite{Chen2022MultiWeatherRemoval}, (c) pixel-level translation~\cite{zhu2017unpaired}, and (d) low-light enhancement~\cite{ma2022toward}. Images processed by these methods either fail to sufficiently enhance image quality or introduce artifacts, affecting semantic prediction accuracy. (e) Our method, AllWeather-Net, effectively enhances color and texture detail while preserving most of the original image information, achieving the best performance.}
        \label{fig:intro2}
\end{figure}

In contrast to weather artifacts removal, pixel-level image translation approaches transform challenging weather situations into clear, sunny-day image styles. Regardless, these methodologies mainly focus only on specific individual conditions, such as rain~\cite{kwak2021adverse} or nighttime scenarios~\cite{anoosheh_night--day_2019}. In addition, the model may alter irrelevant pixels or areas and introduce unwanted changes, leading to visual discrepancies and negatively impacting the performance of downstream tasks. Likewise, low-light enhancement aims to improve the visibility and quality of images captured in low-light conditions. This involves enhancing the brightness, contrast, and details of dark images due to insufficient lighting; however, this technique can mistakenly brighten already well-lit areas, leading to overexposure in weather conditions like snow, as shown in \cref{fig:intro2}.

We aim to improve image quality and clarity by adjusting image attributes and enhancing texture under four distinct adverse conditions, all within a unified framework. Subsequently, we seek to improve semantic segmentation performance. To achieve this goal, we need to consider several critical factors:

\emph{Firstly}, while a unified network is cost-effective, weather variability introduces instability in the learning process. Therefore, it is crucial to identify a stable and invariant signal that can guide the network's learning, ensuring consistent performance across all conditions. \emph{Secondly}, unfavorable conditions differently impact various regions within a captured image. For example, in foggy scenes, distant objects are more blurred than nearby ones due to light scattering and attenuation. In addition, adverse weather conditions tend to preserve larger patterns in images while diminishing the clarity of finer details. So, it is essential to focus on both the overall enhancement and the intricate recovery of texture details. This motivates us to design a network architecture that is contextually aware and sensitive to variations in texture. \textit{Lastly}, employing a pair-to-pair training strategy can improve performance, yet finding perfectly matched pairs in autonomous driving scenes is challenging due to inaccurate GPS pairing and environmental variations. Alternatively, we consider adopting a strategy that utilizes roughly aligned images for more robust discrimination during training when exactly matched pairs are unavailable.

To address these challenges, we propose a novel architecture, namely AllWeather-Net, and our contributions can be summarized as follows:
\begin{itemize}
  \item We are the first to introduce a unified image enhancement method to address image quality degradation under adverse weather and low-light conditions, including snow, rain, fog, and nighttime.

  \item To achieve robust image enhancement across various adverse conditions, we introduce a Scaled Illumination-aware Attention Mechanism (SIAM) that directs a balanced learning process towards different road elements irrespective of changes in weather and scenes.
  
  \item To achieve both overall image consistency and detailed enhancement, we design a novel architecture that enhances input images by conducting discrimination tasks at three hierarchical levels of semantic patches: scene, objects, and texture.

\end{itemize}

\section{Related Work}
In this section, we review the image processing techniques for adverse weather conditions and low-light environments.

\vspace{1mm}\noindent{\bf Weather effect removal.} Current methods for removing visual artifacts, including raindrops, fog particles, and snowflakes, utilize processes such as deraining~\cite{yang2019joint,wang2022uformer}, dehazing~\cite{berman2016non,zhang2019joint,zhang2021hierarchical} and desnowing~\cite{liu2018desnownet,zhang2021deep,wang2023smartassign}. Recently, a unified bad weather removal network was proposed in~\cite{li_all_nodate-1}. In~\cite{jose_valanarasu_transweather_2022}, the researcher simplifies this architecture with a single encoder-single decoder network. To reduce the computational cost,~\cite{Chen2022MultiWeatherRemoval} proposed a knowledge transfer mechanism via teacher-student architecture. 

\vspace{1mm}\noindent{\bf Pixel-level translation} transforms the visual representation and convert adverse weather conditions into scenes resembling sunny, daytime environments. It involves a direct modification of the image pixels, altering the fundamental appearance and context of the scene.  CycleGAN~\cite{zhu2017unpaired} introduced a cycle-consistency loss for unsupervised translation between the source and target domains. CUT~\cite{park2020cut} uses contrastive learning to ensure content preservation and style transfer. Santa~\cite{xie2023unpaired} proposes an approach to find the shortest path between source and target images without paired information.

\vspace{1mm}\noindent{\bf Low-light enhancement} aims to adjust attributes of an image, such as lighting and color balance, to enhance the visual appearance in low-light conditions. Traditional methods utilize histogram equalization~\cite{abdullah2007dynamic} and Retinex~\cite{jobson1997properties} to perform low-light image enhancement. Recent deep learning approaches proposed end-to-end framework~\cite{ma2022toward,fu2023learning,guo_zero-reference_2020}. Compared to traditional methods, these frameworks demonstrate the capability of enhancing the quality of images captured in low-light conditions.

\vspace{1mm}\noindent{\bf Limitation of Existing Works.} 
Removing weather-related unfavourable effects typically targets minor disturbances such as snowflakes or raindrops in the image. However, merely eliminating these atmospheric particles is insufficient, as the primary cause of image quality degradation often stems from alterations in colors and texture details, which significantly contribute to domain shifts. This limitation also applies to pixel-level translation, which frequently introduces unwanted artifacts, thereby reducing the overall image quality and constraining their applicability in safety-critical scenarios. Similarly, low-light enhancement techniques, while focusing on improving visibility under low-light conditions, do not adequately address the challenges posed by adverse weather conditions.

\section{Proposed Method}
Our proposed method uses a generative model for generating image enhancement masks based on the original input image. We introduce a scaled illumination-aware attention mechanism (SIAM) within a unified framework to focus learning on road elements regardless of weather condition. Additionally, our hierarchical framework performs discrimination at multiple semantic levels, ensuring consistent and detailed enhancement. To further ensure precise enhancement, we utilize a ranked adaptive window pairing strategy for accurate discrimination.

\subsection{Enhancement Pipeline}
\begin{figure}[tbp]
    \centering
    \begin{subfigure}[b]{0.47\textwidth} 
        \centering
        \includegraphics[width=\textwidth]{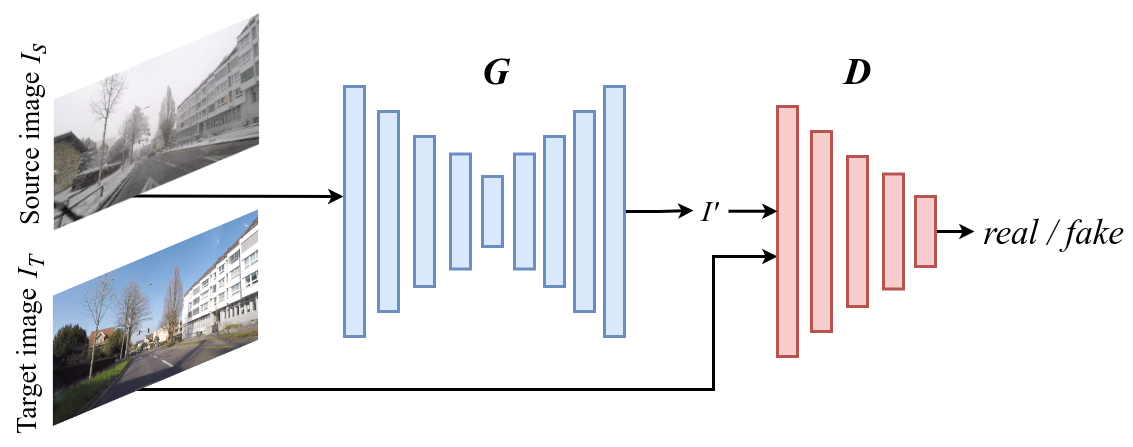}
        \caption{Pixel-level translation}
        \label{fig:pixel_trans}
    \end{subfigure}
    \hspace{0.5cm} 
    \begin{subfigure}[b]{0.47\textwidth} 
        \centering
        \includegraphics[width=\textwidth]{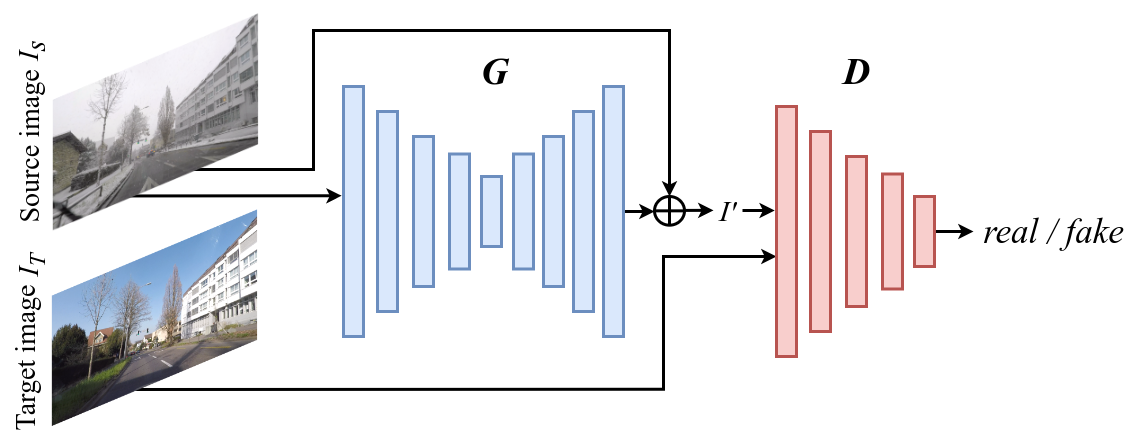}
        \caption{Image enhancement}
        \label{fig:gan_en}
    \end{subfigure}
        \caption{Comparison of pixel-level translation and image enhancement process.}
\end{figure}

Our image enhancement method involves two networks: a generator and a discriminator, which are trained simultaneously through adversarial training to enhance image quality. Unlike pixel-level translation (~\cref{fig:pixel_trans}), where the generator takes the source image $I_S$ and directly outputs translation results to mimic the style of the target image, our image enhancement generates intermediate results that are then combined with the original image $I_S$ to produce final enhancement results. As illustrated in~\cref{fig:gan_en}, the process of generating enhanced image $I'$ can be formulated as:
\begin{align}
I' = G(I_S) + I_S.
\end{align}
\begin{figure}[tbp]
    \centering
    \begin{minipage}{\textwidth} 
        \begin{subfigure}[b]{\textwidth}
            \includegraphics[width=\textwidth]{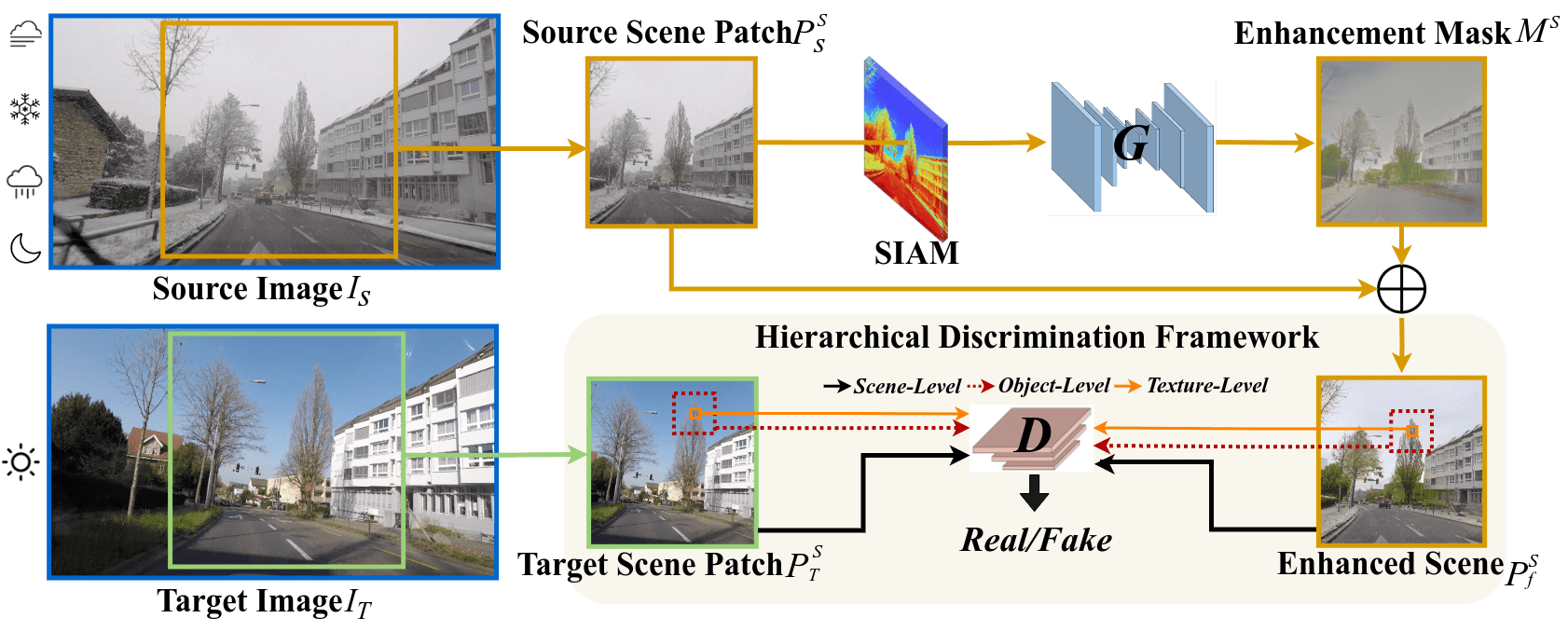} 
            \vspace{-5mm}
            \label{fig:archi1}
        \end{subfigure}
        \caption{Overview of AllWeather-Net architecture. SIAM: Scaled Illumination-aware Attention Mechanism. AllWeather-Net can enhance images across all adverse conditions (\textit{e.g.,} fog, snow, rain, nighttime) with the help of the proposed SIAM and Hierarchical Discrimination Framework.}
        \label{fig:weathernet}
    \end{minipage}
\end{figure}
Pixel-level translation often suffers from generating unwanted artifacts, which can be attributed to the large search space during the training process. By conditioning the output on the input image, the image enhancement model can effectively reduce the search space for the generated result through residual learning. This method ensures that the model's outputs are contextually relevant to the original image, thereby significantly reducing the likelihood of producing unwanted artifacts and improving the quality of the generated images.

In our model (~\cref{fig:weathernet}), we initially cropped the same area from the paired source images $I_S$ and target image $I_T$ as input. The cropped source scene patch $P^s_S$ is processed through a scaled illumination-aware attention mechanism and the generator to produce an enhancement mask $M^s$. This mask is then added to generate the final enhanced results $P^s_f$ and evaluated by different discriminators according to its scale. 

\subsection{Scaled Illumination-aware Attention Mechanism (SIAM)}
\label{sec:3}

Training a unified network for image enhancement across different adverse conditions is cost-effective yet challenging. Each condition uniquely alters the scene's visibility, color, and texture, complicating the learning of consistent information across different scenarios. This variability can significantly degrade the model's capacity to effectively enhance images, especially when adverse conditions heavily obscure scene details. Given these complexities, the importance of a condition-invariant signal in guiding the learning process is paramount. Such a signal should guide the model to learn critical aspects of the scene regardless of weather or lighting conditions. 

Drawing inspiration from previous work~\cite{jiang2021enlightengan} targeting low-light conditions, we consider using illumination as a guiding cue. However, the naive approach of employing illumination intensity as attention tends to overemphasize areas of low illumination while neglecting well-lit regions. This can result in inadequate focus on pixel regions obscured by snow or fog, which often appear brighter due to higher illumination levels. This discrepancy can lead to suboptimal learning outcomes, as crucial details in these areas may not receive sufficient attention, resulting in inconsistent enhancements in the generated images.
\begin{figure}[tbp]
    \centering
    \begin{minipage}{0.5\textwidth} 
        \begin{subfigure}[b]{\textwidth}
            \includegraphics[width=\textwidth]{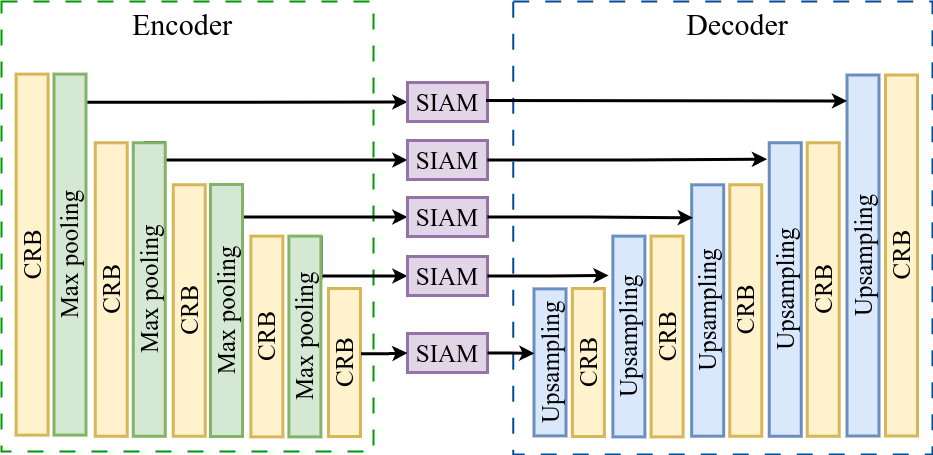} 
            \vspace{-5mm}
            \label{fig:acdc_intensity}
        \end{subfigure}
        \caption{Scaled illumination-aware attention mechanism in the generator.}
        \label{fig:avg_illu_vis}
    \end{minipage}
    \hfill
    \centering
    \begin{minipage}{0.39\textwidth} 
        \includegraphics[width=\textwidth]{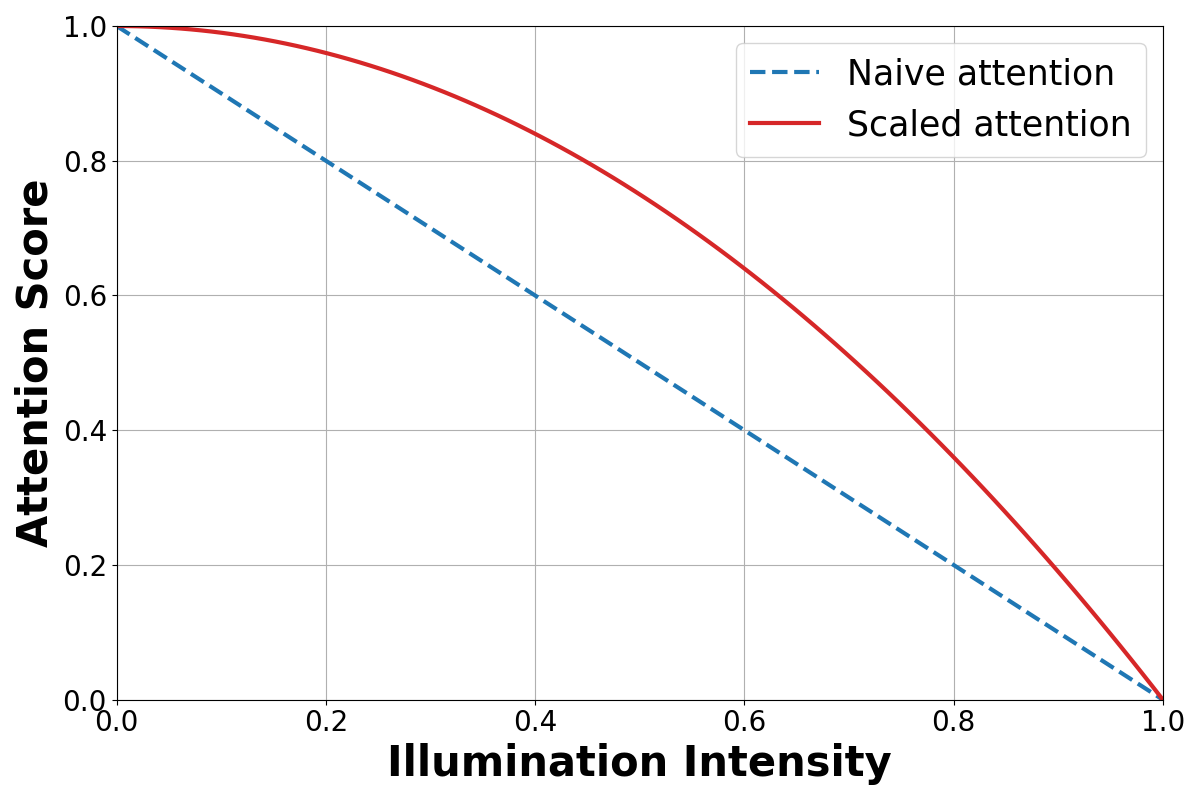} 
        \vspace{-5mm}
        \caption{Attention scores by illumination: naive vs. scaled attention.}
        \label{fig:att_equa_vis}
    \end{minipage}
\end{figure}

To direct a balanced learning for different road elements, rather than merely focusing on dark regions, we propose the scaled illumination-aware attention mechanism (SIAM) to allocate reasonable attention based on illumination intensity. Let \( I_{ij} \) and \( \text{Att}_{ij} \) denote the illumination intensity and the naive illumination attention value at the given pixel location (\(i,j\)), the scaled illumination attention \( \text{S}\_\text{Att}_{ij} \) can be formulated as follows:
\begin{align}
\text{Att}_{ij} & = 1 - I_{ij}, \\
\text{S\_Att}_{ij} & = -\text{Att}_{ij} \cdot (\text{Att}_{ij} - 2).
\end{align}

\begin{figure}[tb]
    \centering

        \includegraphics[width=0.9\textwidth]{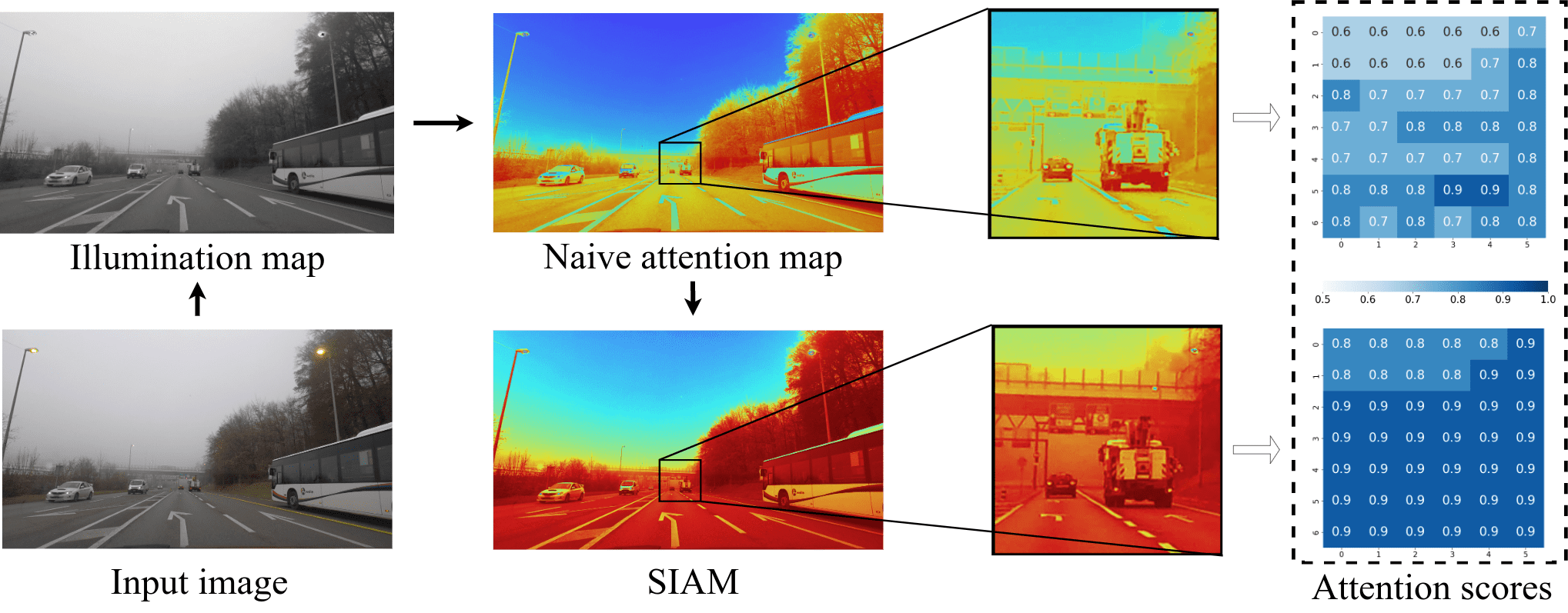} 
        \caption{The flow of generating attention map with SIAM and the comparison between naive attention and SIAM at the image and patch levels. Note that a higher attention score indicates that the model is paying more attention to such an area. This observation suggests that the presented SIAM, compared to naive attention mechanisms, is more adapt at focusing on areas containing  road elements.
 }
        \label{fig:att_fog}
\end{figure}

In the network, the SIAM will guide the learning throughout the generator shown in~\cref{fig:avg_illu_vis}. 
As shown in~\cref{fig:att_equa_vis}, the scaled illumination-aware attention exhibits higher scores for low illumination and maintains consistently high attention levels across the input range for low illumination. This design demonstrates heightened sensitivity towards regions with low illumination while ensuring that high-illumination areas are allocated reasonable focus. With the implementation of the scaled attention, our model prioritizes objects in the distance obscured by fog particles with high illumination (\cref{fig:att_fog}).

\subsection{Hierarchical Discrimination Framework}
\label{sec:1}
\subsubsection{Hierarchical Discrimination.} From the perspective of discrimination, employing a single discriminator generates unrealistic colors, while the global-local structure~\cite{jiang2021enlightengan} has limited performance in providing fine-grained textural details.

To address this issue, we propose our hierarchical discrimination architecture with scene-, object-, and texture-level patches/discriminators.
\begin{figure}[htbp]
        \centering
        \includegraphics[width=\textwidth]{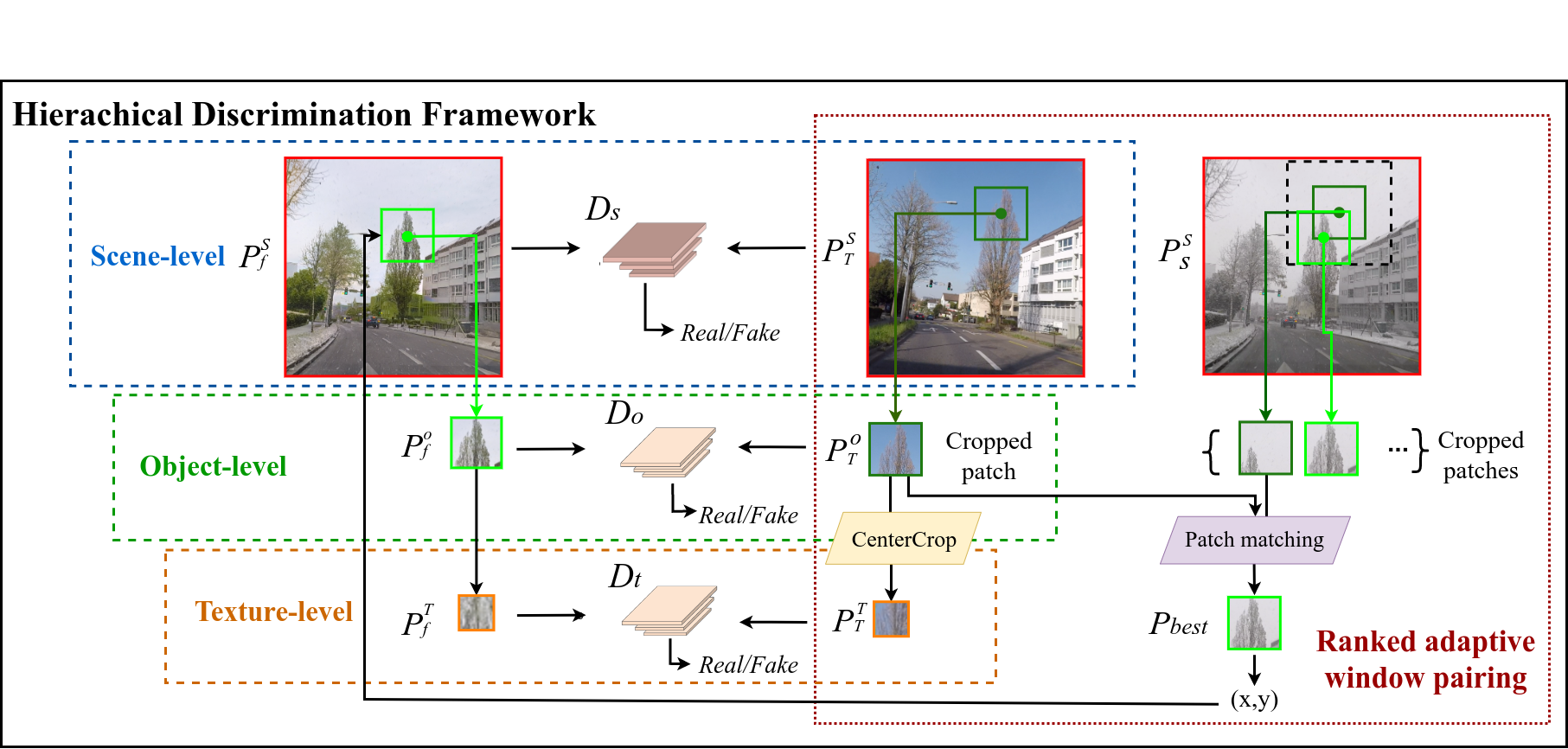}
        \caption{The details of hierarchical discrimination framework.}
        \label{fig:hie_dis}
\end{figure}

The scene-level discriminator $D_s$ assesses a randomly cropped scene patch $P^s$ from the original image $I_S$, as shown in~\cref{fig:hie_dis}. Its primary purpose is to evaluate the overall coherence and realism of the generated scene. Next, we crop object patches $P^s$ with $1/4$ of the size of the scene-level patch for the object-level discriminator $D_o$. This selective cropping allows for the exploration of intricate image details that the scene-level discriminator might overlook due to its broader perspective. Lastly, we derive the texture-level patch $P^t$ by $1/4$ center cropping from each object-level patch $P^o$. The texture-level discriminator $D_t$ represents the highest level of inspection in our discriminator hierarchy, focusing on examining fine details and texture quality of the generated image.

By incorporating the mentioned three different levels of patches and discriminators, our model differentiates between the broad scales and fine levels of detail. This enables generated images that are not only realistic in overall appearance but also exhibit improved textural fidelity. 

\subsubsection{Ranked Adaptive Window Pairing.}
\label{sec:2}
Scene-level discrimination is accurate due to similar information in paired images. However, accuracy drops at the object level, where patch pairs often show significant information shifts due to more apparent changes in perspective, leading to suboptimal outcomes.

To address the above mentioned issue, we utilize an adaptive window with a ranked score to identify the object-level patches that are most closely aligned. We begin by cropping object-level patches $P^o$ at the same location from scene-level patches $P^s$. Moreover, we then define a fixed search area $A$ with a width of $w$ and a height of $h$. Within this area, we deploy a dynamic window $W$, of size $z \times z$, to traverse the defined area with a stride of $s$, thereby generating object-level patch candidates. These candidate patches are subsequently compared to the corresponding target patch using a ranked pairing score to determine the best match of $P^o_S$ and $P^o_T$. Finally, we center crop the top-matched object pairs to obtain patches $P^t_S$ and $P^t_T$ for texture-level discrimination. Let $(x, y)$ represent the top-left coordinate of the matched source object patch. $N=\frac{x + w - z}{s}$ represents the horizontal steps required, while $M = \frac{x + h - z}{s}$ denotes the vertical steps needed to traverse the search area with the dynamic window. The formulation of the pairing score $F$ is defined as:

\begin{equation}
F(P^o_{Sc}, P^o_T) = \sum_{i=0}^{N} \sum_{j=0}^{M} \left| I_{P^o_T}(x, y) - I_{P^o_{Sc}}(x + i \cdot s, y + j \cdot s) \right|,
\label{eq:lpair}
\end{equation}
where $P^o_{Sc}$ denotes the \( c \)-th candidate patch within the search window area and $I$ represents the RGB values of the patch. The variables \( i \) and \( j \) signify the horizontal and vertical offsets within 
the search window, respectively. The best-matched patch, denoted as \( P_{\text{best}} \), is then determined:
\begin{equation}
\mathit{P}_{\text{best}} = \arg\min_{P \in \text{W}} S(P^o_{Sc}, P^o_T).
\end{equation}

By identifying the location of $P_{\text{best}}$ in the source scene, we can locate its counterpart in the corresponding generated scene patch. 

\subsection{Loss Function}
We utilize a relativistic approach~\cite{jolicoeur2018relativistic} that compares the realism between real and generated images. We employ LSGAN~\cite{Mao_2017_ICCV} loss for direct assessment of the realism of the object- and texture-level discrimination. The scene-level losses for the discriminator and generator are given below:
\begin{equation}
L^{\text{s}}_{D} = \mathbb{E}_{P^s_r \sim P_{\text{real}}} \left[(D_{s}(P^s_r, P^s_f) - 1)^2\right] 
+ \mathbb{E}_{P^s_f \sim P_{\text{fake}}} \left[D_{R}(P^s_f, P^s_r)^2\right],
\label{eq:scenedisloss}
\end{equation}
\begin{equation}
L^{\text{s}}_{G} = \mathbb{E}_{P^s_f \sim P_{\text{fake}}} \left[(D_{s}(P^s_f, P^s_r) - 1)^2\right] 
+ \mathbb{E}_{P^s_r \sim P_{\text{real}}} \left[D_{R}(P^s_r, P^s_f)^2\right],
\label{eq:scenegenloss}
\end{equation}

\noindent where $D_{s}$ represents the relativistic discriminator, $P^s_{r}$ and $P^s_{f}$ denote the real and generated fake scene patch.
$\mathbb{E}_{P^s_r \sim P_{\text{real}}}$ and
$\mathbb{E}_{P^s_f \sim P_{\text{fake}}}$ represent expectations over the real and fake  data distributions. For object- and texture-level loss, The discriminator and generator for  losses \(P^x \in \{P^o, P^t\}\) are given by:

\begin{equation}
L^x_{D} = \mathbb{E}_{P^x_r \sim P^x_{\text{real}}} [(D_x(P^x_r) - 1)^2] + \mathbb{E}_{P^x_f \sim P^x_{\text{fake}}} [(D_x(P^x_f) - 0)^2],
\end{equation}

\begin{equation}
L^x_{G} = \mathbb{E}_{P^x_f \sim P^x_{\text{fake}}} [(D_x(P^x_f) - 1)^2].
\end{equation}

Here, \(P^x_r\) and \(P^x_f\) denote real and fake patches of type \(x\), respectively. \(D_x\) represents the discriminator for the patch type \(x\), and $\mathbb{E}_{P^x_f \sim P_{\text{fake}}}$ and $\mathbb{E}_{P^x_r \sim P_{\text{real}}}$ are the same meaning as for scene patch. Consider the set $\mathcal{L} = \{s, o, t\}$ corresponding to the types of losses and use $\lambda_1$, $\lambda_2$, and $\lambda_3$ to control each loss contribution to balance loss; the total training loss can be written as:
\[
\mathit{Total\ Loss} = \sum_{\ell \in \mathcal{L}} \Lambda_{\ell} \cdot (L^{\ell}_{G} + L^{\ell}_{D}),
\]
where $\Lambda_{s} = \lambda_1$, $\Lambda_{o} = \lambda_2$, and $\Lambda_{t} = \lambda_3$.

\section{Experiments}
We conduct our experiments on image enhancement and evaluate the outcomes from two perspectives: image quality and semantic segmentation. Both aspects are assessed qualitatively and quantitatively.

\begin{figure}[tbp]
  \centering
  \begin{subfigure}{\linewidth}
    \centering\includegraphics[width=\linewidth]{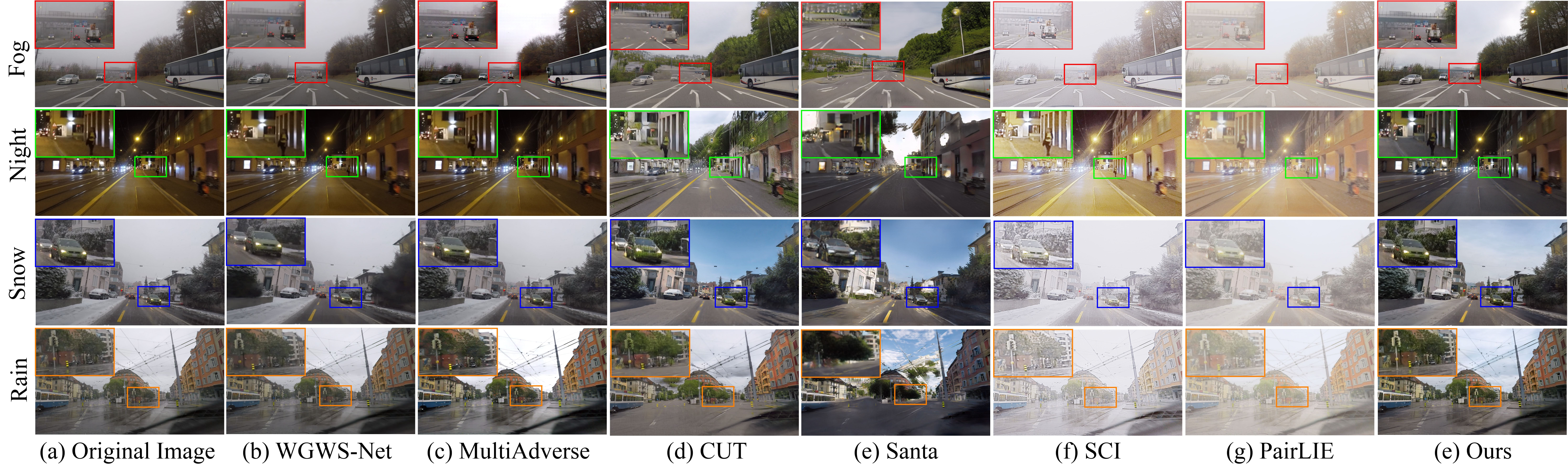}
  \end{subfigure}
  \caption{Comparison with other image processing methods regarding weather effect removal, pixel-level translation, and low-light enhancement, using zoomed-in red regions to highlight visual distinctions.}
  \label{fig:Comp}

\end{figure}
\subsection{Dataset}
For image enhancement model training, we use 1,600 images from the ACDC~\cite{sakaridis_acdc_2021}, evenly distributed among snow, rain, night, and fog conditions, and 2,416 nighttime images from the Dark Zurich~\cite{dai_curriculum_2019}. 
For the evaluation of semantic segmentation, our model enhances images from both the ACDC and the Dark Zurich validation set, which are subsequently tested using a pre-trained PSPNet~\cite{zhao_pyramid_2017} model.
To demonstrate the generalization capabilities of our model, we apply it to the test datasets of Foggy Zurich~\cite{sakaridis2018model} and Nighttime driving~\cite{dai2018dark}.

\subsection{Comparisons}
We evaluate AllWeather-Net against three distinct approaches: (a) weather effect removal, (b) pixel-level translation, and (c) low-light enhancement. We employ the Peak Signal-to-Noise Ratio (PSNR) and Structural Similarity Index Measure (SSIM) to assess enhancement quality. We utilize the Natural Image Quality Evaluator (NIQE) to evaluate the image naturalness. Additionally, we consider the similarity between the improved and reference daytime images employing the Learned Perceptual Image Patch Similarity (LPIPS). We adopt the Mean Intersection over Union (mIoU) metric to evaluate semantic segmentation performance.

\vspace{1mm}\noindent{\bf Image quality.}
In~\cref{fig:Comp}, the first column represents the input images captured under different adverse conditions while the subsequent columns are the enhanced results by different models. We observe that weather effect removal methods such as WGWS-Net~\cite{zhu2023Weather} and MultiAdverse~\cite{Chen2022MultiWeatherRemoval} excel in eliminating weather-related artifacts but fall short in achieving a sunny daytime appearance. Pixel-level translation methods, including  CUT~\cite{park2020cut} and Santa~\cite{xie2023unpaired} can effectively transform images from adverse conditions to daytime-like settings, yet these models struggle with visual consistency in nighttime scenes and may inadvertently remove less prominent objects, such as a car in the distance, in foggy conditions. Low-light enhancement-based method SCI~\cite{ma2022toward} and PairLIE~\cite{fu2023learning} can enhance lighting and brightness in night scenes but over-expose in well-lit conditions, \eg fog and snow. In contrast, our method demonstrates the most significant improvements in brightness and contrast across various scenes. It delivers the most realistic and clear daytime representation under diverse adverse conditions, significantly enhancing visibility without introducing artifacts or excessive noise. Additionally, it outperforms other models in color correction and detail enhancement, offering a more comprehensive solution. 

As shown in~\cref{tab:acdc} and~\cref{tab:dark}, our method achieves the most natural image enhancement outcomes with the lowest NIQE score and excels in converting images to the daytime domain, indicated by the lowest LPIPS scores for nighttime scenes.

\begin{table}[tbp]
\large
    \centering
    \caption{Qualitative evaluation of image quality and semantic segmentation performance on ACDC dataset.}
    \label{tab:model_performance_by_type}
    \resizebox{0.82\textwidth}{!}{%
        \begin{tabular}{c|l*{5}{|c}}
            \hline
            \multirow{2}{*}{\textbf{Method Type}} & \multirow{2}{*}{\textbf{Models}} & \multicolumn{5}{c}{\textbf{Metrics}} \\
            \cline{3-7}
            & & \textbf{SSIM$\uparrow$} & \textbf{PSNR$\uparrow$} & \textbf{NIQE$\downarrow$} & \textbf{LPIPS$\downarrow$} & \textbf{mIoU$\uparrow$} \\
            \hline
            \multirow{2}{*}{Weather effect removal} & WGWS-Net~\cite{zhu2023Weather} & 0.3916 & 11.4812 & 0.1480 & 0.4740 & 36.4 \\
            & MultiAdverse~\cite{Chen2022MultiWeatherRemoval} & 0.3822 & 10.9135 & 0.1782 & 0.4752 & 37.0 \\
            \hline
            \multirow{3}{*}{Pixel-level translation} & CycleGAN~\cite{zhu2017unpaired} & 0.3981 & \textbf{12.2493} & 0.1578 & 0.4655 & 33.6 \\
            & CUT~\cite{park2020cut} & 0.3776 & 12.1043 & 0.1668 & 0.4833 & 29.3 \\
            & Santa~\cite{xie2023unpaired} & 0.3920 & 12.0863 & 0.1374 & 0.4770 & 25.7 \\
            \hline
            \multirow{4}{*}{Low-light enhancement} & EnlightenGAN~\cite{jiang2021enlightengan} & 0.3905 & 11.8725 & 0.1651 & 0.4649 & 37.2 \\
            & Zero-DCE~\cite{guo_zero-reference_2020} & 0.3160 & 10.6015 & 0.2726 & \textbf{0.4428} & 32.4 \\
            & SCI~\cite{ma2022toward} & 0.3239 & 8.5693 & 0.2600 & 0.5149 & 24.4 \\
            & PairLIE~\cite{fu2023learning} & 0.3577 & 8.9133 & 0.1564 & 0.4598 & 28.3 \\
            \hline
            All weather enhancement & \textbf{Ours} & \textbf{0.3983} & 11.6618 & \textbf{0.1257} & 0.4619 & \textbf{38.2} \\
            \hline
        \end{tabular}
     \label{tab:acdc}}
\end{table}

\begin{table}[tbp]
    \centering
    \caption{Qualitative evaluation of image quality and semantic segmentation performance on Dark Zurich dataset.}
    \label{tab:model_performance_overall}
    \resizebox{0.6\textwidth}{!}{%
        \begin{tabular}{l*{5}{|c}}
            \hline
            \multirow{2}{*}{\textbf{Models}} & \multicolumn{5}{c}{\textbf{Metrics}} \\
            \cline{2-6}
            &  \textbf{SSIM$\uparrow$} & \textbf{PSNR$\uparrow$} & \textbf{NIQE$\downarrow$} & \textbf{LPIPS$\downarrow$} & \textbf{mIoU$\uparrow$} \\
            \hline
            EnlightenGAN~\cite{jiang2021enlightengan} & 0.3791 & \textbf{10.2460} & 0.3186 & 0.4766 & 10.8 \\
            Zero-DCE~\cite{guo_zero-reference_2020} & 0.3324 & 8.6448 & 0.3425 & 0.5109 & 10.6 \\
            SCI~\cite{ma2022toward} & 0.3519 & 8.5735 & 0.2554 & 0.5115 & 11.0 \\
            PairLIE~\cite{fu2023learning} & 0.3827 & 9.1442 & 0.2083 & 0.4844 & 7.1 \\
            \hline 
            \textbf{Ours} & \textbf{0.3849} & 9.6121 & \textbf{0.1850} & \textbf{0.4589} & \textbf{17.6} \\
            \hline
        \end{tabular}
     \label{tab:dark}}
\end{table}

\begin{figure*}[tbp]
  \centering
  \begin{subfigure}{\linewidth}
    \centering
    \includegraphics[width=\linewidth, height=2.8in, keepaspectratio]{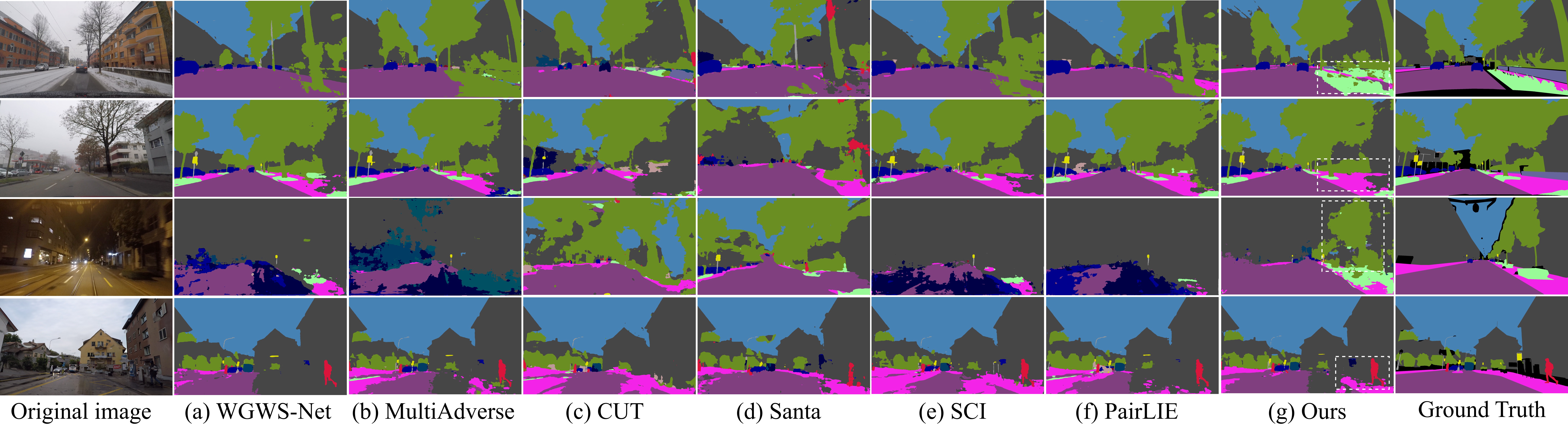}
  \end{subfigure}
  \caption{Semantic segmentation results in comparison with other state-of-the-art methods of weather effect removal, pixel-level translation, and low-light enhancement, using zoomed-in white regions to highlight visual distinctions.}
  \label{fig:sem_pred}
\end{figure*}

\vspace{1mm}\noindent{\bf Semantic segmentation.} The effectiveness of our image enhancement model for semantic segmentation is assessed by performing a direct evaluation using the pre-trained PSPNet model~\cite{zhao_pyramid_2017}. We apply the model to datasets enhanced by our model as well as those enhanced by others. As shown in last column in~\cref{tab:acdc} and~\cref{tab:dark}, our method demonstrates superior performance in both adverse weather and nighttime scenes. This indicates that our model can enhance the performance of semantic segmentation models by significantly improving visual quality and visibility compared to other image processing models. In the visualization results shown in \cref{fig:sem_pred}, our method improves the detail recognition of road elements such as trees, grass, and pedestrians in all conditions.

\vspace{1mm}
\begin{figure*}[tbp] 
  \centering
    \includegraphics[width=\linewidth]{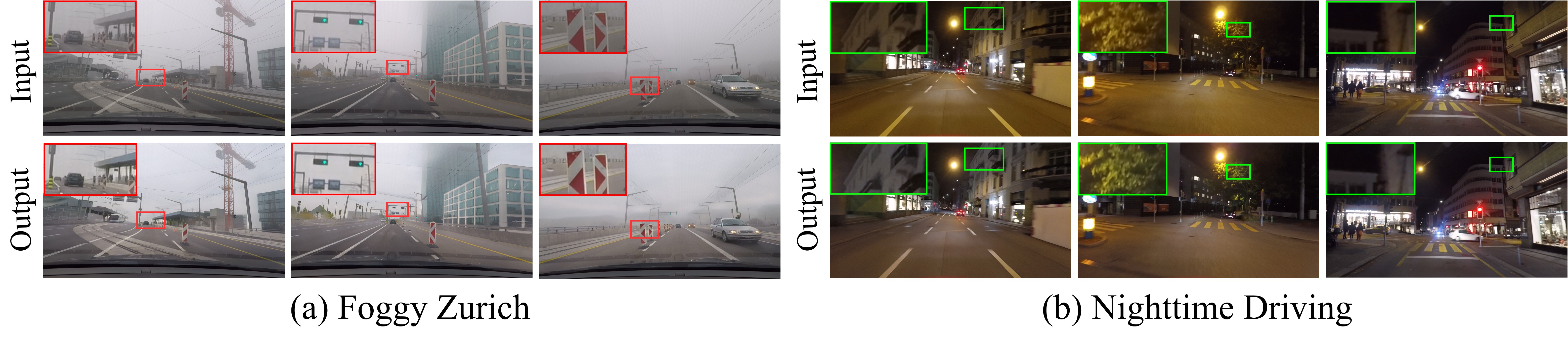}
    \vspace{-4mm}
    \caption{The generalization performance of our model on the Foggy Zurich and Nighttime Driving datasets. The red and green box corresponds to the zoomed-in patches.}
    \label{fig:generalization}
\end{figure*}

\vspace{1mm}\noindent{\bf Generalization to unseen datasets.}
We evaluate our trained model's performance in scenarios not seen during training using the Foggy Zurich and Nighttime Driving datasets. The results in ~\cref{fig:generalization} show that our model enhances clarity of cars, traffic lights, and road signs in the Foggy Zurich dataset, and corrects the yellowish glow on buildings and trees in the Nighttime Driving dataset, restoring their true colors and visibility. The mIoU comparison (\cref{tab:gen_iou}) shows improvements of 1.8\% and 3.9\% respectively, highlighting the remarkable generalization capability of our model and demonstrating its ability to enhance semantic segmentation performance in unseen domains without re-training.
\begin{table}[H]
\centering
\caption{MIoU scores tested on pretrained PSPNet \cite{zhao_pyramid_2017} for original and enhanced versions of Foggy Zurich and Nighttime Driving test datasets.}
\label{table:miou}
\adjustbox{width=0.5\textwidth}{
\begin{tabular}{@{\hspace{0.2cm}}l|@{\hspace{0.5cm}}c@{\hspace{0.5cm}}c@{\hspace{0.2cm}}}
\toprule
\textbf{Datasets} & \textbf{Original} & \textbf{Enhanced} \\ \midrule
Foggy Zurich \cite{sakaridis2018model} & 26.3 & \textbf{28.1} \\
Nighttime Driving \cite{dai2018dark} & 23.0 & \textbf{26.9} \\ \bottomrule
\end{tabular}
\label{tab:gen_iou}
}
\end{table}
\subsection{Ablation Studies}
To demonstrate the impact of each component in our method, we conduct ablation experiments with a focus on image quality improvement. These experiments examine different levels of discrimination, Ranked Adaptive Window Pairing (RAWP), and the Scaled Illumination Attention mechanism (SIAM).
\begin{table}[t]
    \centering
    \caption{Qualitative comparison of model components using SSIM. A higher SSIM value indicates better image generation quality as it signifies greater similarity to a clear daytime image in terms of structure, luminance, and contrast.}
    \footnotesize
    \begin{tabularx}{0.6\textwidth}{c|*{5}{>{\centering\arraybackslash}X}|c}
        \hline
        \multirow{2}{*}{\textbf{Models}} & \multicolumn{5}{c|}{\textbf{Components}} & \multirow{2}{*}{\textbf{SSIM} $\uparrow$} \\
        & \(D_{\text{s}}\) & \(D_{\text{o}}\) & \(D_{\text{t}}\) & RWAP & SIA &  \\\hline
        $M_1$ & \checkmark & & & & & 0.3864 \\
        $M_2$ & \checkmark & \checkmark & & & & 0.3879 \\
        $M_3$ & \checkmark & \checkmark & \checkmark & & & 0.3910 \\
        $M_4$ & \checkmark & \checkmark & \checkmark & \checkmark & & 0.3922 \\
        $M_5$ & \checkmark & \checkmark & \checkmark & \checkmark & \checkmark & 0.3983 \\ \hline
    \end{tabularx}
    \label{tab:abla}
\end{table}

In table~\cref{tab:abla}, we observe enhanced image quality, as indicated by higher SSIM values, with the incremental inclusion of discriminators: scene discriminator \(D_{\text{s}}\), object discriminator \(D_{\text{o}}\), and texture discriminator \(D_{\text{t}}\). The introduction of RWAP further increases SSIM, indicating that the model learns finer details in local patches through pair-to-pair training. Subsequently, incorporating attention further elevates SSIM, demonstrating the effectiveness of attention in enhancing image quality. 
\begin{figure}[t]
  \centering
  \begin{minipage}[t]{0.56\linewidth}
    \centering
    \includegraphics[width=\linewidth]{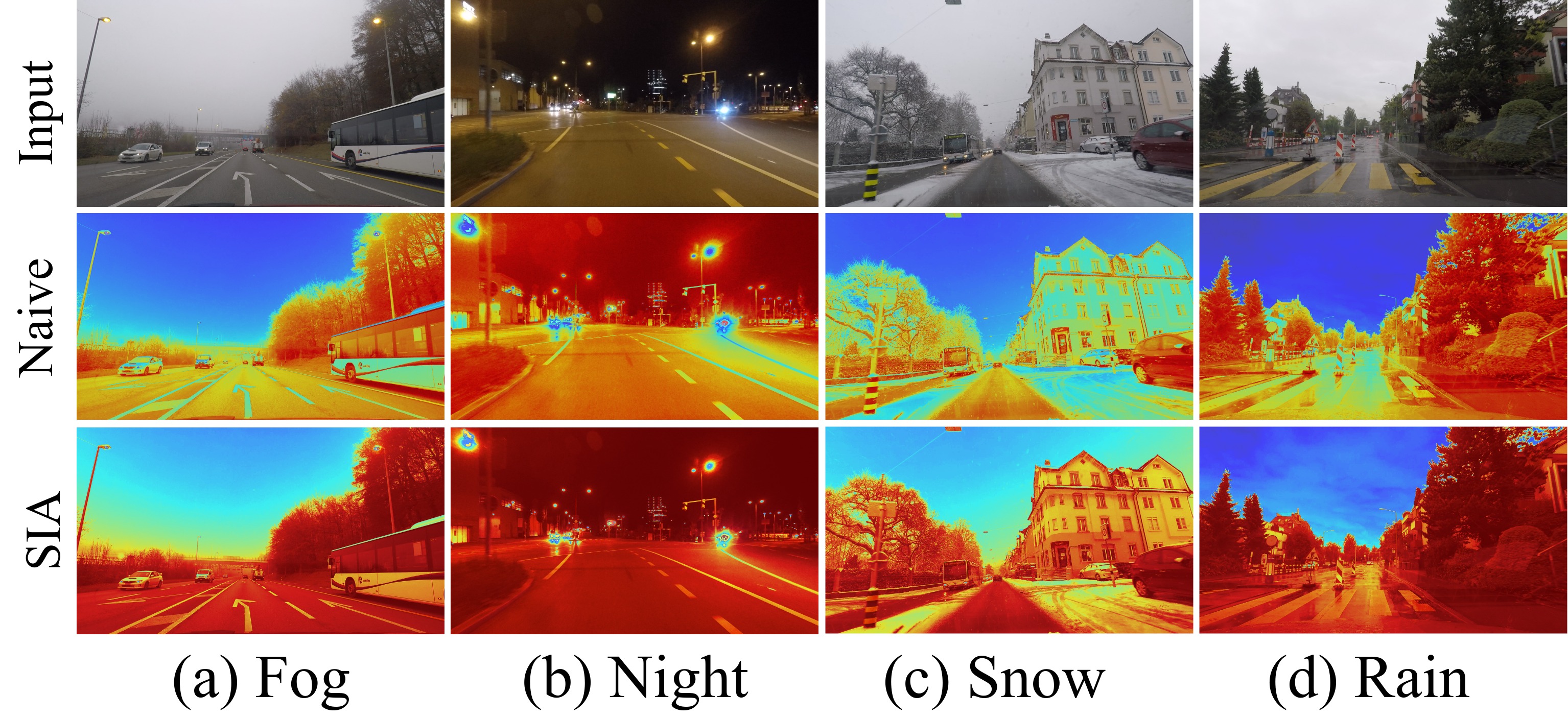}
    \vspace{-6mm}
    \caption{Naive attention and SIAM maps for various input adverse condition images. Darker regions indicate higher attention scores.}
    \label{fig:att_a}
  \end{minipage}
  \hspace{5pt} 
  \begin{minipage}[t]{0.4\linewidth}
    \centering
    \includegraphics[width=\linewidth]{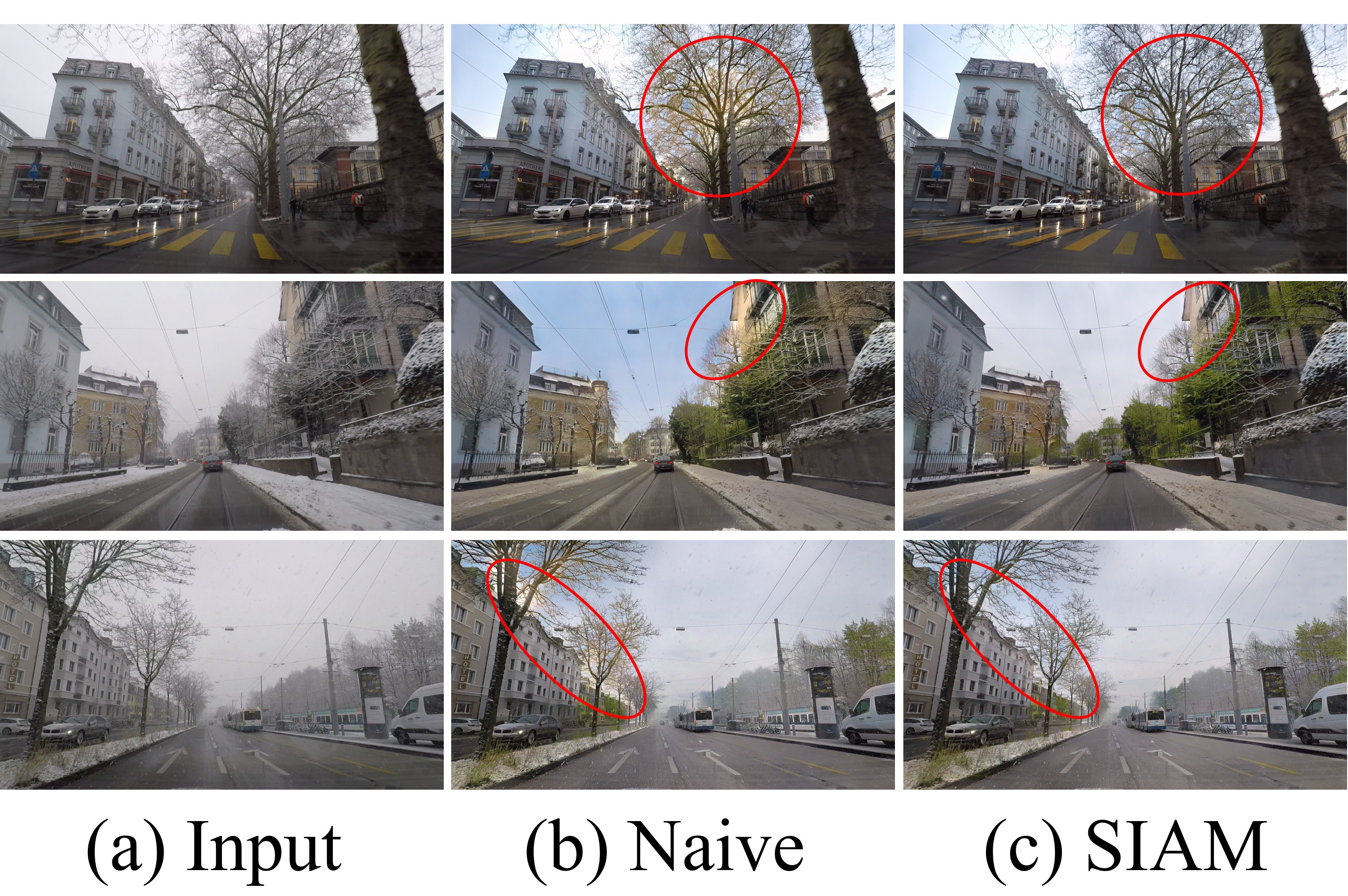}
    \vspace{-6mm}

    \caption{Results generated by models trained with naive attention and scaled attention.}
    \label{fig:att_b}
  \end{minipage}
  \label{fig:ablation_sia}
\end{figure}
For various adverse conditions, the scaled illumination attention mechanism effectively focuses on road elements, as demonstrated in~\cref{fig:att_a}. The enhanced image quality in~\cref{fig:att_b} indicates that scaled attention addresses uneven lighting issues by allocating attention appropriately, particularly to areas overlaid with high illumination. This highlights the capability of the scaled attention mechanism to focus on both low and high-illumination regions within road elements and adapt to all adverse conditions.

\section{Conclusions}
In this work, we introduced AllWeatherNet, a unified framework designed to enhance image quality under various adverse conditions such as snow, rain, fog, and nighttime. Our objective was to develop a singular model capable of simultaneously addressing these four conditions without introducing artifacts that degrade image quality. The model can adjust lighting, brightness, and color in images 
in both adverse and normal weather conditions, transforming them into clear, daytime-like visuals. We implemented a hierarchical framework to recover color and texture details, along with a ranked adaptive window pair-to-pair training strategy to boost performance. We also developed a scaled-illumination attention mechanism to direct the learning process towards low and high-illumination areas, making it adaptable to different adverse scenarios. We performed semantic segmentation experiments using our enhanced dataset and observed notable improvements. Additionally, the model demonstrated exceptional generalization capability across a range of datasets without requiring re-training. 



\end{document}